\documentclass{tlp}

\usepackage{mathptmx}
\usepackage{amsmath}
\usepackage{algorithm}
\usepackage{graphicx}
\usepackage{algpseudocode}

\definecolor{linkcolor}{RGB}{52,59,144}
\usepackage[hidelinks, colorlinks=true, allcolors=linkcolor]{hyperref}
\usepackage{url}

\usepackage{listings}

\usepackage{comment}
\usepackage{cprotect}

\usepackage{fancyvrb}
\usepackage{soul}
\usepackage{url}
\usepackage{booktabs}
\usepackage{subcaption}

\usepackage{textcomp}
\usepackage[dvipsnames]{xcolor}
\definecolor{PrologPredicate}{RGB}{0,0,200}
\definecolor{PrologVar}      {RGB}{145,032,039}
\definecolor{PrologComment}  {RGB}{169,082,044}
\definecolor{PrologOther}    {rgb}{0.2,0.2,0.2}
\definecolor{PrologString}   {RGB}{070,120,200}
\usepackage{relsize}
\usepackage{listings} 

\lstdefinestyle{MyInline}
{
  basicstyle = \ttfamily\color{PrologOther},
  breaklines = true,
  breakatwhitespace=true,
  upquote = true,
}
\lstdefinestyle{MySCASP}
{
  keywords = {},
  breaklines = false,
  upquote = true,
  basicstyle = \relsize{-.5}\ttfamily\color{PrologPredicate},
  basewidth = 0.44em,
  moredelim = {*[s][\color{black!40!PrologPredicate}]{\#pred}{.}},
  moredelim = {*[s][\color{black!40!PrologPredicate}]{\#show}{.}},
  moredelim = {*[s][\color{black!40!PrologPredicate}]{\#hide}{.}},
  moredelim = {*[s][\color{PrologVar}]{(}{)}},
  moredelim = {*[s][\color{PrologString}]{'}{'}},
  moredelim = {*[s][\color{PrologOther}]{:-}{.}},
  commentstyle = \mdseries\color{PrologComment},
  morecomment=[l]\%,
}
\lstdefinestyle{Themes}
{
  keywords = {},
  breaklines = true,
  breakatwhitespace=true,
  basicstyle = \relsize{-.5}\ttfamily\color{PrologString},
  basewidth = 0.44em,
}
\usepackage[textwidth=0.12\textwidth,textsize=scriptsize,backgroundcolor=none,
]{todonotes}
\usepackage{enumitem}

\begin{document}

\jnlPage{X}{Y}
\jnlDoiYr{2024}
\doival{10.1017/xxxxx}

%\title[Socialbots with LLMs and Automated Reasoning]{Building Reliable Interactive Socialbots with LLMs and Automated Reasoning}
\title[A Socialbot built using LLMs and s(CASP)]{A Reliable Common-Sense Reasoning Socialbot\\ Built Using LLMs and 
Goal-Directed ASP.}

\begin{authgrp}
  \author{%
    \sn{Zeng}, \gn{Yankai$^1$} \hspace{1cm}
    \sn{Rajashekharan}, \gn{Abhiramon$^1$} \hspace{1cm}
    \sn{Basu}, \gn{Kinjal$^2$} \\
    \sn{Wang}, \gn{Huaduo$^1$}  \hspace{1cm}
    \sn{Arias}, \gn{Joaqu\'in$^3$}  \hspace{1cm}
    \sn{Gupta}, \gn{Gopal$^1$}
  }
  \vspace*{1em}
  \affiliation{$^1$University of Texas at Dallas, Richardson, USA}
  \affiliation{$^2$IBM Research}
  \affiliation{$^3$CETINIA, Universidad Ray Juan Carlos, Madrid, Spain}
  %\email{}
  %$\{$yankai.zeng, abhiramon.rajashekharan, kinjal.basu, huaduo.wang, gupta$\}$@utdallas.edu}
\end{authgrp}

\history{\sub{xx xx xxxx;} \rev{xx xx xxxx;} \acc{xx xx xxxx}}

\maketitle

\begin{abstract}
%Today, conversational social bots can now be quickly constructed using Large Language Models (LLMs). However, relying solely on LLMs often leads to conversations that are out of control and that deviate from their original purpose. Due to the reliance on pattern matching rather than on knowledge, it is challenging for LLMs to stay grounded and up-to-date. This inevitably leads to hallucinatory responses. Due to reliance on pattern matching,  LLMs also struggle with linking various topics cohesively, resulting in mediocrity and tedium when they have to deviate from the main topic to keep the conversation interesting.

%To mitigate these challenges, we propose to use an Answer Set Programming (ASP)-based reasoner to ensure certainty in a Socialbot's responses. Such systems ensure consistency and accuracy and regulate conversation dynamics very much how a human would manage it. In our proposed Socialbot framework, the LLM component handles natural language parsing, converting sentences into predicates, and generating appropriate responses based on reasoning over predefined topics. Concurrently, the ASP reasoner enhances certainty through logical reasoning, guides topic transitions, and offers recommendations for a more coherent conversation flow. Our main contribution in the paper is to show that effective automated socialbots can be built using LLMs as a semantic parser and ASP as a backend reasoner. Building socialbots that can hold interesting conversations and that do not stray off-topic is an open problem that we solve.

The development of large language models (LLMs), such as GPT, has enabled the construction of several socialbots, like ChatGPT, that are receiving a lot of attention for their ability to simulate a human conversation.
  % Problem (and related work limitations)
However, the conversation is not guided by a goal and is hard to control. In addition, because LLMs rely more on pattern recognition than deductive reasoning, they can give confusing answers and have difficulty integrating multiple topics into a cohesive response. These limitations often lead the LLM to deviate from the main topic to keep the conversation interesting.
% Proposal
We propose AutoCompanion, a socialbot that uses an LLM model to translate natural language into predicates (and vice versa) and employs commonsense reasoning based on Answer Set Programming (ASP) to hold a social conversation with a human. In particular, we rely on s(CASP), a goal-directed implementation of ASP as the backend.
This paper presents the framework design and how an LLM is used to parse user messages and generate a response from the s(CASP) engine output.
% Evaluation/validation
To validate our proposal, we describe (real) conversations in which the chatbot's goal is to keep the user entertained by talking about movies and books, and s(CASP) ensures (i) correctness of answers,
%certainty in the answers, 
(ii) coherence (and precision) during the conversation---which it dynamically regulates to achieve its specific purpose---and (iii) no deviation from the main topic.
% AR - What is certainty?

\end{abstract}


\section{Introduction}
\label{sec:introduction}

Conversational agents are designed to understand dialogs and generate meaningful responses to communicate with humans. After the popularity of ChatGPT, with its surprising performance and powerful conversational ability, commercial \textit{Large Language Models} (LLMs) for general NLP tasks such as GPT-4 by \cite{gpt4}, Gemini from \cite{gemini}, etc., sprung up and brought the generative AI as a solution to the public view. These LLMs work quite well in content generation tasks, but their deficiency in 
%reviewer update 5
knowledge-and-reasoning-oriented tasks are well-established by now, according to \cite{chatgpt-critic}. 
These models themselves cannot tell whether the text they generate is based on facts or made-up stories, and they cannot always follow the given data and rules strictly and sometimes even modify the data at will, also called \textit{hallucination}. The reasoning that these LLMs appear to perform is also at a very shallow level. These are serious flaws that make the LLMs unsuitable for fact-based conversations such as providing correct information to a user. 
The reason for this flaw is that LLMs generate text that is purely based on a pattern-matching mechanism, and consequently have no understanding of the meaning of sentences and thus 
%reviewer update 5
a safe and sound way of reasoning is not guaranteed, as stated by \cite{chatgpt-critic}.
In contrast, humans understand the meaning of sentences and then use their reasoning capabilities to check for consistency and take further action. Thus, to make the machine-generated response reliable and consistent, our socialbot needs to follow a similar approach. 

Following the above insights, in this paper, we report on developing an elaborate conversational socialbot that can understand human dialog and respond properly according to human expectations. The goal of this socialbot is to be able to engage in multiple rounds of social conversations with users about their interests, understand the context of what the user is saying like a human, and be able to initiate new topics or make relevant recommendations at the right time to control the pace of the chat. We currently focus our topics of interest on discussions of movies, books, and related people, and use the STAR framework by \cite{star}, which has been successfully applied in the task-oriented chatbot domain (e.g., AutoConcierge from \cite{autoconcierge}), to build a social conversational bot called 
%reviewer update 3
AutoCompanion.\footnote{Now available at \url{https://github.com/Sambour/Social-bot/}}
AutoCompanion uses an LLM to interact with the user and to translate between natural language dialogs and knowledge represented in predicates. It further uses an ASP system for reasoning over this knowledge. 
After the user's input is parsed into predicates by the LLM, the ASP reasoner uses reasoning (and a bit of randomness) to decide whether to continue with the current topic of conversation, move on to a related topic, or recommend books or movies that are currently on sale based on the user's preferences. Questions asked by the user are also answered. Subsequently, the instructions for the next step from the reasoner are passed on to another LLM in the form of an open-ended free-for-all on that topic, and the generated sentence is provided to the user as a reply.
%All these abilities are achieved by \textit{reasoning} through the structured predicate expression captured from the user's input. These predicates are generated through the use of LLMs. 
We believe that the use of LLM should be controlled to avoid its misuse in fact-based domains and that the best way to utilize LLM is to use it only as an interface for parsing and presenting knowledge. %Following this intuition, we designed and built our AutoCompanion that 
Our AutoCompanion outperforms LLM-only socialbots in reliability, scalability, controllability, creativity, and many other aspects.
%  AR - controllability might be a better word for manipulability?

Our contribution in this paper are as follows. \textbf{First}, We developed a general socialbot framework that allows the bot to control the conversation flow and react reliably according to the user input, attributed to leveraging both the LLM and s(CASP), as described in Section \ref{sec:design}. Section \ref{sec:ontology} shows the predicates we extract from natural language sentences, Section \ref{sec:reasoning} explains how the reasoner works to add control and consistency to the system, and Section \ref{response} provides the details to the response generation stage. \textbf{Second}, We built the AutoCompanion based on the above framework with knowldge about books and movies, as described in Section \ref{sec:implement}. Section \ref{sec:sample} displays a short conversation between our AutoCompanion and the user.
{Finally}, this paper also discusses the usage and performance (in Section \ref{sec:usage} and \ref{sec:discussion}) of LLMs on the similar tasks.

\section{Background and Related Work}

This section introduces s(CASP), the reasoner used to control the conversations. It also presents state-of-the-art Chatbots and discusses their limitations.

\subsection{Answer Set Programming and s(CASP)}
% Modify from the same part in the STAR paper
 
Answer Set Programming (ASP), according to \cite{cacm-asp} and  \cite{gelfond2014knowledge}, is a logic programming paradigm suited for knowledge representation and reasoning that facilitates commonsense reasoning.  
The s(CASP) system proposed by \cite{scasp}, is an answer set programming system that supports predicates, constraints over non-ground variables, uninterpreted functions, and, most importantly, a top-down, query-driven execution strategy.
These features make it possible to return answers with non-ground variables (possibly including constraints among them) and compute partial models by returning only the fragment of a stable model that is necessary to support the answer.% to a given query.
%
%The s(CASP) system supports constructive negation based on a disequality constraint solver, and unlike Prolog's negation as failure and ASP's
%default negation, %
%{\tt not p(X)} can return bindings for \texttt{X} on success, i.e., bindings for which the call \texttt{p(X)} would have failed. 
%Additionally, s(CASP) system's interface with a constraint solver (over reals) allows for sound non-monotonic reasoning with constraints (useful for solving algebra problems in one of the NLU applications we discuss later).

Complex commonsense knowledge can be represented in ASP and the s(CASP) query-driven predicate ASP system can be used for querying it, as shown in the work done by \cite{chef} and \cite{murder-trial}. Commonsense knowledge can be emulated using (i) default rules, (ii) integrity constraints, and (iii) multiple possible worlds, according to \cite{gelfond2014knowledge} and \cite{gupta-csr}. Default rules are used for jumping to a conclusion in the absence of exceptions, e.g., a bird normally flies, unless it's a penguin. 

%Thus, if we are told that Tweety is a bird, we jump to the conclusion that Tweety flies. Later, if we are told that Tweety is a penguin, we withdraw the conclusion that Tweety can fly. Default rules with exceptions represent an elaboration-tolerant way of representing knowledge~\cite{gelfondkahl}. 

\begin{lstlisting}[style=MySCASP]
flies(X) :- bird(X), not abnormal_bird(X).
abnormal_bird(X) :- penguin(X).
\end{lstlisting}

%{\small \tt
%flies(X) :- bird(X), not abnormal\_bird(X).
%
%abnormal\_bird(X) :- penguin(X).}  

\noindent Integrity constraints allow us to express impossible situations and invariants. For example, a person cannot be dead and alive at the same time.

\begin{lstlisting}[style=MySCASP]
false :- person(X), dead(X), alive(X).
\end{lstlisting}

\noindent Finally, multiple possible worlds allow us to construct alternative universes that may have some parts common but other parts inconsistent. For example, the cartoon world of children's books has a lot in common with the real world (e.g., birds can fly in both worlds), yet in the former birds can talk like humans but in the latter they cannot. 
%
%GG: paragraph below can be removed 
%Default rules are used to model a bulk of our commonsense knowledge. Integrity constraints help in checking the consistency of the information extracted. Multiple possible worlds allow us to perform assumption-based (or abductive) reasoning. % (for example, knowing that ``Alice loves Bob'', we could assume that either Bob also loves Alice or he does not). 

A large number of commonsense reasoning applications have already been developed using ASP and the s(CASP) system: \cite{logical-english,chef,murder-trial}. %In the three applications reported in this paper, we have kept the commonsense reasoning component simple, as our main goal is to illustrate our framework for combining LLMs and ASP to develop NLU applications that are explainable and reliable. The goal-directed s(CASP) ASP system is crucial for the AutoConcierge system. It is used to perform commonsense reasoning resembling a human concierge: to conclude, to check for inconsistencies in the knowledge derived from the user's dialogs, etc. 
Justification for each response can also be given as the s(CASP) system can generate justifications for successful queries as proof trees as shown by \cite{scasp-justification}.

\subsection{State-of-the-art Chatbots and Their Limitations}

Conversational agents (chatbots) have been an active area of research for a long time.
%Before LLM-driven chatbots came out, many chatbots had been deployed for businesses, especially e-commerce websites. However, they are deficient, and quickly handover to a human operator after one or two rounds of dialogs. Other commercial chatbots such as Amazon Alexa, Siri, and Google Assistant are more of information retrieval devices rather than actual effective chatbots that can have a stateful conversation.
%The task-oriented chatbots are designed with a closed set of instructions (ontology) and a mapping method matching user commands to the ontology. 
Rule-based or finite-state-based systems, like Eliza by \cite{eliza}, Chat-80 by \cite{chat80}, and PARRY by \cite{parry}, encode the mapping of user commands to ontology using rules and state transitions.
The Conversational Knowledge Template (CKT) approach proposed by \cite{ckt} enables the system to control the dialog flow and change topics. 
Here, the main challenge is parsing the dialogs and correctly extracting the knowledge conveyed in the dialogs. %In recent years, neural models have been employed for the Dialogue State Tracing (DST) task \cite{NBT,sst,dqfd,soloist,pptod,autoconcierge}.

Until recently, transformer-based Large Language Models, pre-trained on an enormous quantity of well-annotated data, have been applied to general NLP tasks. With the advent of Large Language Models, the paradigm changed from pre-training and fine-tuning  (\cite{pre_trained_transformers}) to teaching a language model any arbitrary task using just a few demonstrations, called \textit{in-context learning}, a method of \textit{prompt engineering}. \cite{gpt3} introduced an LLM called GPT-3 containing approximately 175 billion parameters that have been trained on a massive corpus of filtered online text, on which the well-known ChatGPT is based. %of \cite{chatgpt} is based.
GPT-3 and its successor GPT-4 can perform competitively on several tasks such as question-answering, semantic parsing, and machine translation, as noted by \cite{gpt4}. However, such LLMs lack the ability of mathematical reasoning and find it hard to overcome the hallucination brought from the training data, as shown by \cite{gpt3-scope,chain,chatgpt-critic}. 
%After ChatGPT opened a new chapter towards generalized AI, the recent line of related transformer-based large language models to build social chatbots are used for commercial. %Some focus on miniaturization and localization, such as Mistral 7B \cite{mistral7b}. 
% AR - commenting the line below
% These models are based purely on pattern matching and thus suffer from similar disadvantages discussed earlier. 

Retrieval Augmented Generation (RAG) by \cite{rag} is proposed and widely used to mitigate the deficiencies mentioned above by retrieving the relevant materials using similarity matching of content embedded as vectors by a transformer-based model. % and then uses the retrieved material to generate the fact-based answer, which solves to some extent the problem of searching in the context of vast amounts of knowledge. 
Recent efforts by \cite{ragchat1,ragchat2}, are trying to leverage RAG for building chatbots, but none of them engages an explicit reasoning system.

Finally, Our AutoCompanion is an extension of our previous work developing NLU systems based on commonsense reasoning (\cite{ckt,discasp,autoconcierge}). Our group has been dedicated to building socialbots, specifically addressing the \cite{alexa} Alexa Socialbot Challenge for years. GPT-4 with in-context learning as a semantic parser leads to a significant advantage over our previous socialbots and helped this framework succeed, as we report in this paper. 
%in this non-task-oriented conversational domain.
% AR - non-task-oriented conversational domain might be unclear

\section{Socialbot Design}\label{sec:design}

The AutoCompanion is designed to model the human thought process during conversations. When humans hear a sentence, they parse it to extract its meaning and represent the meaning in their minds as knowledge. Humans will check for consistency and correctness of this knowledge using additional (commonsense) knowledge, that also resides in their mind, to find any gaps in the captured knowledge wrt their goal. If human reasoning reveals gaps in this knowledge, they may ask further questions to cover this gap. Once the knowledge is (reasonably) complete, humans will draw a conclusion. This conclusion---that also represents knowledge---will be translated into a natural language response. Therefore, similar to humans, when AutoCompanion receives a user sentence, it will also process it using three stages:
\textrm{First}, it will parse the natural language input to knowledge, which is done by LLMs reliably.
\textrm{Next}, it will check the consistency and correctness of the input knowledge and use the knowledge extracted from the dialog, in conjunction with commonsense knowledge it already possesses, to reason out the next action. The ASP reasoning system is used to implement this capability.
\textrm{Finally}, the knowledge representing the next step (conclusion) is converted into a natural language response by another invocation of the LLM and communicated. 

This mechanism is embodied and known as the STAR framework by \cite{star}, which leverages the advantages of LLMs and ASP systems by combining them systematically. It parses the semantics of the text sentences to generate the predicates using LLMs such as GPT 3.5 or GPT 4. After that, it sends the predicates to the ASP system to get reliable answers through reasoning. This framework was previously applied to task-oriented chatbots such as AutoConcierge by \cite{autoconcierge} and has proved to work well.
% reviewer update 10
Indeed, LLMs do not parse the sentences to knowledge with 100\% accuracy, but LLMs have proven to be the most effective semantic parsers by far. Prior to LLMs, \cite{aaai21} used parsers such as the Stanford CoreNLP parser and the spaCy parser, and those methods were largely ineffective in extracting semantics. Besides, the wrong predicate generated by LLMs can often be checked by additional logic written in ASP.

Unlike task-oriented chatbots, socialbots are not set to complete a task. Hence, they can chat without a goal, until whenever the user likes to end the conversation. This requires AutoCompanion to switch the chat topic actively based on the current and previous conversation. \cite{psy} suggests that a good chatbot should satisfy the user's sense of security, inclusion, and gratification. With this in mind, AutoCompanion is designed to contain the below functions.

\begin{enumerate}
\item AutoCompanion should be able to carry a conversation on the current topic to a deep level and express a refined perspective based on a true understanding of the topic. For example, when the user is talking about the plot of the movie Titanic, the AutoCompanion can have an in-depth discussion about this plot: either the lines, actors' performance, or even comment on it. For example, if the user mentions a famous line in a movie, AutoCompanion can talk about its influence on the plot, etc.

\item AutoCompanion should be able to jump from one topic to another with some common features. Again, taking Titanic as an example, AutoCompanion is capable of fluidly changing the topic from Titanic to Catch Me If You Can, as Leonardo DiCaprio acted in both movies.

\item AutoCompanion should be able to maintain consistency, typically for opinions. This indicates that, once AutoCompanion expresses a point of view, no matter whether positive or negative, it should remain the same for the subsequent time. It can also follow the user's idea if the user insists on a contradicting view.

\item AutoCompanion should be able to answer the user's question accurately. When the user's question exceeds its knowledge, it should answer that it does not know.

\item AutoCompanion should be able to recommend the current popular topics the user might like. For example, if there is one in-theater movie that meets all the preferences the user mentioned previously, AutoCompanion should recommend this movie to the user.
\end{enumerate}

\begin{figure*}[tb]
    \centering
    \includegraphics[width=\linewidth]{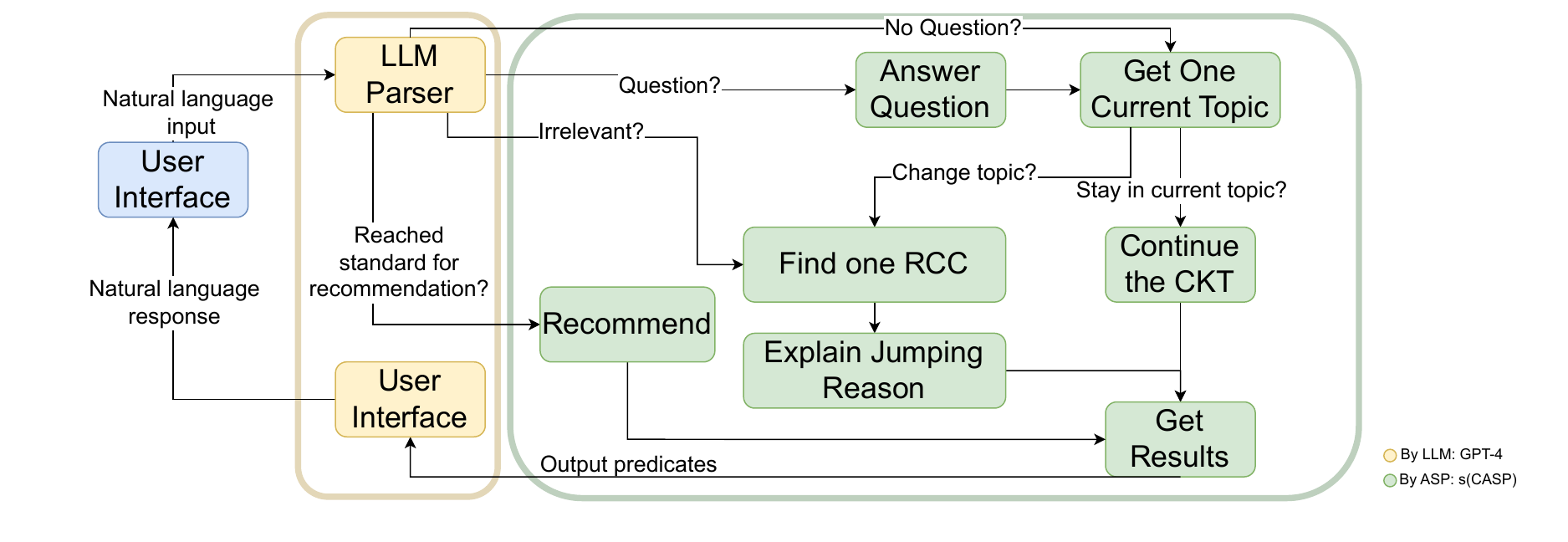}
    % reviewer update 2
    \caption{AutoCompanion Architecture. The yellow-colored boxes are handled by GPT-4 in Python, and the green-colored  by  s(CASP). The two parts interact using Python subprocess calls.}
    \label{fig:arch}
%\vspace{-0.15in}
\end{figure*}

To satisfy these functions, AutoCompanion is designed in the structure as shown in Figure \ref{fig:arch}.
%
% reviewer update 1
From the perspective of conversation, our AutoCompanion is designed to be evoked by the user. However, if the user does not know what to talk about, they can simply say ``Hello.", or ``Hello, let us talk about movies.", and AutoCompanion will guide the user to continue the conversation.

\subsection{LLM Parser and Ontology}\label{sec:ontology}
AutoCompanion leverages LLMs such as the GPT series to parse the natural language sentences to knowledge. The prompts are given to LLMs including the knowledge ontology, which is represented in the predicates used in the reasoning part, and several examples covering the use cases.

\noindent \textbf{Topic}. The topic describes a subject to talk about in a social conversation. A socialbot may cover many different topics when chatting with the users, such as movies, books, sports, pets, etc., while a certain movie, book, sport, or pet discussed is called the topic instance. AutoCompanion will mainly cover movies, books, and relevant people.

\noindent \textbf{Property}. A property indicates an aspect of the topic instance to be potentially discussed. The list of properties is fixed for each topic and covers all possible directions that can be used.

\noindent\textbf{Theme}. A theme is a set of predicates centralized on one property of a topic instance. It typically contains one talk predicate: ``talk(Topic, Instance\_name, Property).'', one attitude predicate: ``attitude(positive/negative/ask/argue).'', and one content predicate: ``content(Property, Detailed\_content).''. If the user asks a question, the predicate question: ``question(What\_is\_asked)'' is also captured. For instance, when the user compliments the scene of the movie Titanic where the giant ship is sinking, the predicates extracted from the user would be:
\begin{lstlisting}[style=Themes]
talk(movie, `Titanic', scene). content(scene, sinking). attitude(positive).
\end{lstlisting}

\noindent\textbf{Preference}. The user preference is also extracted to be used later for recommendation. The preference is presented by the prefer predicate: ``prefer(Topic, Property, Preference).''. 

In the parsing stage, the prompts are given to extract all themes and preferences a user input is about, where different themes and preferences are separated by `\#\#\#'. LLMs also extract the exception predicates such as `quit' and `irrelevant' at this stage. The prompt is given following the below template (only part of the prompts are shown due to space limitation):

{\small\tt Now you are an expert about books and movies. Now you should use your 

~~profession to extract predicates from sentence input...

The predicates are below:

talk(Topic, Name, Property) \%\% The topic can be movie or book or people, ...

Examples:

Sentence 1 -> theme 1 \#\#\# theme 2 ...

Input sentence ->}

\subsection{Topic Control}\label{sec:reasoning}
We leverage an ASP reasoner to guide and switch between the topics. Predicates from the LLM parser are separated into themes and preferences that are recorded and updated, and one theme is randomly picked to respond with. The remaining themes will be used in later rounds where nothing is captured from the user's input. Then, the ASP reasoner chooses whether to remain in the same topic instance or switch to a relevant one by the below strategy: if the instance is first mentioned, the reasoner maintains it for one more round; otherwise, a random function is applied to decide whether to continue the current instance, the current property, or start a relevant one. The maintaining and shifting can be well described by two separate mechanisms, \textit{CKT} and \textit{RCC}.

\smallskip\noindent 
\textbf{Conversational Knowledge Template (CKT)} by \cite{ckt} is used when the ASP reasoner continues the current topic instance. It maintains a state with a list of properties that can be discussed for each topic instance. Once the reasoner chooses to go deeper into one instance using other properties, the CKT will be triggered to pick up one property that has not been discussed yet for this topic instance. On the other hand, the CKT can also choose to remain on the current property, which will lead the LLM response generator (see Section \ref{response}) to tell something new about this property (e.g., depict another plot, etc.). %Typically, suppose the user mentions something about one property of the instance, indicating that the user wants to go on it, the response will remain the same property, and the sentences generated by LLM will cater to the user's preference.

CKT also controls AutoCompanion's attitude about an instance and its properties. Four kinds of attitudes are used in AutoCompanion: positive, negative, ask, and acknowledge. For a new topic instance or a new property of the current instance, AutoCompanion would randomly pick one attitude from positive or negative. In contrast, for an existing instance property, it will obey the previous attitude, which is based on the assumption that generally humans have a consistent attitude towards one property of a specific instance and AutoCompanion simulates this human feature. The attitude ``ask'' is involved in the above two cases of attitude selection, in which case, instead of expressing ``personal'' views, AutoCompanion asks the user about their attitude toward it. The attitude ``acknowledge'', indicating maintaining the current property, is only applied if the user argues on one property of the topic instance.

\smallskip\noindent 
\textbf{Relevant Consistent Concept (RCC)} selector, proposed by \cite{discasp}, will be invoked if the ASP reasoner decides to switch to a relevant topic instance. In our AutoCompanion, an RCC is a new topic instance that shares some common properties with the current instance. For example, the RCCs of a movie include movies of the same genre, or by the same director, %movies starring the same actors, books adapted from the movie, the people involved in the movie production, 
etc. These relations are formatted and listed for reasoner to search the RCCs automatically.

The RCC searching step maps the current topic instance to the RCC relation rules and enumerates all possible RCCs in the current conversation round. Next, if the reasoner chooses to switch to a relevant topic, one of the RCCs will be randomly picked, together with its relation to the source topic. The RCC selecting mechanism guarantees that the original topic instance is related to the target one. It can discover connections that are not obvious.

\smallskip\noindent 
\textbf{Question Answering}: Whenever user asks some questions, AutoCompanion answers them in the next round of replies. The AutoCompanion would first track all current themes that contain a question, and then search through the knowledge base for an 
answer. These used themes are no longer recorded in the unused theme list. When the answer to the question does not lie in the knowledge base, the AutoCompanion would honestly say ``I don't know'' instead of trying to find an answer using its ambiguous knowledge learned from the LLM pre-training, which minimizes hallucinations to a great extent.

\smallskip\noindent
\textbf{Recommendation}: Our AutoCompanion also collects the user preference during the conversation by the ``prefer'' predicates, which are then recorded and updated independently by maintaining a state. Every time the ``prefer'' predicate is extracted, AutoCompanion tracks the latest news of the discussed topics (e.g. the movie in the theater today, or the bestselling books today, etc.) that meets the user's preference. Once the number of matched properties meets a threshold, AutoCompanion will recommend this new topic instance. While recommending a topic, the AutoCompanion would not use the themes, and the extracted themes will be reserved for further conversation. After recommending, if the user does not actively talk about it, AutoCompanion will avoid recommending it again repeatedly. Instead, it goes to the general response mode and waits for the correct conditions for the next recommendation. 

Following this pattern, recommendations for new topic instances will not appear so abrupt but will be smoothly integrated into the conversation, making the conversation more engaging. 

\smallskip \noindent 
\textbf{Response to Irrelevant Dialogs}: The response prevents the chat from being off-topic. It is triggered if the user talks about something irrelevant to the interest-sharing, or the LLM parser cannot extract useful information from user input. Under the irrelevant mode, the AutoCompanion would first reply ``I cannot catch up with you now. Let's go back and talk about ...'' and loop back to the CKT or RCCs selection step by 
% AR - Should it say 'I cannot catch up with you', or 'I don't understand'.
randomly picking an unused theme in the theme list. If the theme list is empty, it will try to start with some famous movies, such as Titanic. The irrelevant mechanism ensures the conversation is always centralized in the prescribed topics, and the AutoCompanion continues chatting (by proposing a relevant topic instance) even if the user exhausts their words about a certain topic instance.

\smallskip\noindent
\textbf{Output Format}: The output of the reasoner contains three parts: \textit{mode}, \textit{answers}, and \textit{reply theme}. The \textit{mode} ranges over ``quit'', ``irrelevant'', ``general'', and ``recommend'', where the former two are tied with the two special predicates, and the latter two define whether to talk about the next theme or make a recommendation. The \textit{answers} part has all the answers to the given questions. In the recommend mode, the \textit{reply theme} refers to the latest topic instance to recommend and the reason to recommend it; while in the general mode, it indicates the theme to talk about in the next round, the attitude toward it, and how it is associated with the former theme if a relevant theme is selected.

\subsection{Response Generation}\label{response}
Due to the lack of detailed information about each movie (e.g. for plots, lines, etc.), book (e.g. for plots, characters, etc.), and person (e.g. for skills, appearance, etc.), our AutoCompanion leaves the generation of these specifics to LLM, controlling only the topic and property for each round. We assume that the LLMs are experts in common topics such as famous movies and books, which makes it possible for us to leverage this advantage of LLMs in constructing AutoCompanion. Once a comprehensive knowledge base is built for these topics, control over the specific content level can also be exerted.

While leveraging the LLMs for content generation, a proper prompt makes the output closer to the expected performance. In our designed prompt, we first describe the conversation situation and the LLM's role and define the response tone and length. Then we provide the LLM with some examples and ask it to generate the content following the format of the examples. The examples always start with a question with the template: 

{\small\tt What are the most interesting [property] for [topic] [instance]? [attitude] ->}
% AR - Maybe saying 'interesting' instead of 'impressive' in the prompt might get more engaging responses from the bot?
% YZ: That's possible

\noindent And LLM would generate an answer for that. For example, If we ask about the value expressed in Batman Begin:

{\small\tt What are the most interesting value expressed for Batman Begins? positive ->}

\noindent The LLM would generate:

{\small\tt  I feel like Batman Begins nails it with the whole inner battle vibe. It's not just about taking down bad guys; it's Bruce Wayne wrestling with his own demons, and that's what makes it hit home. Do you have the same feeling as me?}

\noindent If the LLM cannot find any specific content that supports the given attitude, we allow the LLM to oppose its attitude toward the content, and this will be captured by the topic controller.

The function of the ``attitude'' predicate is to help LLMs generate some alternative perspectives other than the mainstream ones. During the AutoCompanion construction, we found out that LLMs seldom create a view that strays too much from the common views of the public. However, for an ordinary human, not all of their opinions may match the mainstream. Therefore, randomness should be added to the attitude to help LLMs generate ``surprising'' responses.

If the RCC mechanism is invoked, the response generator will also create cohesive sentences that naturally mention the next topic instance to discuss. The template for a cohesive sentence is:

``{\small\tt Because you mentioned [source\_topic\_instance], it makes me think of the [next\_topic] [next\_topic\_instance], since [relation\_of\_them].}''

\noindent Similarly, if the reasoner answers the questions from the user, one of the following templates will be used according to the condition:

``{\small\tt I remembered that the [property] of the [topic] [topic\_instance] is [answer].}''

``{\small\tt Sorry I could not remember the [property] of the [topic] [topic\_instance].}''

\noindent And the recommendation generation adopted the below template:

``{\small\tt Do you know the recent [topic] named [topic\_instance]? Since you like [source\_topic], so you should like it.}''

The reply sentences generated by the above templates are then concatenated and processed by an LLM language modifier to make the replies more coherent, gentler, and more informal.

\section{Implementation}\label{sec:implement}

AutoCompanion is constructed in Python with the s(CASP) goal-directed ASP Reasoning Engine (\cite{scasp}). GPT-4 is chosen as the LLM in both the LLM Parser and Response Generation stages. 

\subsection{Data Collection}
We deployed our AutoCompanion to discuss movies, books, and the people involved.

\smallskip\noindent
\textbf{Movies:} The movie data is from IMDb dataset,\footnote{IMDb dataset is available at \url{https://developer.imdb.com/non-commercial-datasets} and \url{https://cinemagoer.github.io}.} where we collected the information of release year, runtime, rating, countries involved, languages, genres, top 10 casts with their characters, directors, writers, editors, composers, producers, cinematographers, and the plot summary. % Here the plot summary is more like the preview or trailer of the film that only provides the general description of the story background, and thus cannot be used in the content generation. 
The above information is then marked as properties for answering users' questions and linking the movies to related people. 
Other than those above, the properties of the movie topic include plot episodes, scenes, lines, costumes, awards, music, values expressed, characterization, cinematography, technique, actor performance, adaption, and social impact, which are out of the knowledge base but are likely to be mentioned by the users. %, and the users would probably be interested in talking about them, and thus we leave it to LLMs to play freely. Due to the slow response of the web request of the IMDb dataset, 
We store the database locally with a selection of 500 top-rated movies, and the 500 most popular movies since 2010, resulting in 931 movies.

\smallskip\noindent
\textbf{Books:} They were collected from the Users-Books-Dataset,\footnote{Users-Books-Dataset is available at \url{https://data.world/divyanshj/users-books-dataset}.} including 528 books with information about their series, author, rating, language, genres, awards, setting places, characters, and plot description. These are also recorded as properties along with their writing style, symbolism, emotional impact, and social background.

\smallskip\noindent
\textbf{People:} The data of the people associated with the movies are also extracted from the IMDb database and stored locally. The people topic contains properties of birth year, death year, profession, and representative work, whereas the skills, awards, appearance, and personal life require an LLM to generate. We extracted all people linked to the selected movies, resulting in a knowledge base of 5625 data. The authors in the book's knowledge base are also matched to the corresponding movie people.

%An adaption knowledge base is constructed to map the movies to the original novel it is adapted from and used in RCC searching.

Identifying \textbf{in-theater movies} and \textbf{bestsellers in bookstores} is also required for recommendation mode. We leveraged IMDb's website for movie in-theater by providing location information, %\footnote{\url{https://www.imdb.com/showtimes/location/}}. 
and chose USA Today's Best-Selling Booklist\footnote{USA Today's Best-Selling Booklist is available at \url{https://www.usatoday.com/booklist/booklist}.} as our best-selling book data and extract further details from Open Library.\footnote{Open Library is available at \url{https://openlibrary.org/dev/docs/api/search}.} The knowledge from the recommendation knowledge base  is used to match the user preference (the ``prefer'' predicate) in the property of popularity rank, rating, genre, language, located countries, writer, actor, director, etc. %In the LLM parsing step, the property rank and rating can take the following values: ``high'', ``above average'' and ``any'', separately map to the manually set thresholds.

% reviewer update 8
\subsection{ASP Coding}

Below are several coding examples of the ASP implementation of the reasoning part of our AutoCompanion. In this part, we will elaborate on how the CKT and topic switching work in our s(CASP) program.

\smallskip\noindent
\textbf{CKT implementation}. The CKT component chooses what property to discuss under a topic. The code for CKT is shown below:

\begin{lstlisting}[style=MySCASP]
next_attr(Topic, Name, Attr) :- continue_attr, talk(Topic, Name, Attr), 
                                response_attr(Topic, Attr).
next_attr(Topic, Name, Attr) :- response_attr(Topic, Attr), 
                                not neg_next_attr(Topic, Name, Attr).
neg_next_attr(Topic, Name, Attr) :- not next_attr(Topic, Name, Attr).
neg_next_attr(Topic, Name, Attr) :- hist(I, Topic, Name, Attr, Attitude, From).
\end{lstlisting}

The first rule indicates to continue the current property if continue\_attr, which is evoked by a random function, is true.
The second and third rule forms an even loop, making it possible to choose any available property.
The fourth rule defines that the discussed property should not be selected.

\smallskip\noindent
\textbf{Topic switching}. When a topic continues for several rounds, the reasoner may switch to a related topic. This is implemented by the codes below.

\begin{lstlisting}[style=MySCASP]
next_topic(Topic, Name, Attr, 'None', 'None') :- 
        round(I), hist(I, Topic, Name, Attr, Attitude, 'user'), 
        not discussed_before(Topic, Name, Attr).
next_topic(Topic, Name, Attr, 'None', 'None') :- 
        continue_topic, talk(Topic, Name, Attr1), 
        next_attr(Topic, Name, Attr).
next_topic(Topic, Name, Attr, Source, Relation) :- 
        len_rcc(I), random(1, I, N), rcc(N, Topic, Name, Source, Relation), 
        next_attr(Topic, Name, Attr).
\end{lstlisting}

The first rule sets out that a new topic should continue for at least one round. The second rule indicates to continue the current topic if continue\_topic is true.
The third rule describes how to switch the topic to another linked one by a random RCC.

\subsection{Leveraging LLMs}\label{sec:usage}

Besides parsing and response generation, LLMs are also used for other purposes.

\smallskip\noindent
\textbf{Property enumeration}. Before constructing the AutoCompanion, a detailed list of properties is needed for each topic. Although the property list has been equipped with information extracted from the knowledge base, more properties are required to cover as many themes as people might be interested in while chatting. Therefore, an LLM is used to seek a comprehensive list of properties. We used GPT-4 with the following prompt to list the properties.

``{\small\tt When people talk about their interests in [topic] in a social chat, what aspects will they discuss? Please make each point as much different as possible from others. Summarize each point in a short name.}''

\noindent We ran GPT-4 for 5 rounds to ensure comprehensiveness and collected all the answers it gave, concatenating them with the information from the knowledge base to form the property list.

\smallskip\noindent
\textbf{Name correction}. LLM is also used to correct the names of the topic instances extracted from user input. Due to the LLM parser only serving to turn the sentences into predicates, sometimes even when the user inputs the incorrect topic instance name, the LLM parser copies the mistakes as they are. Also, some instance names have multiple spellings, resulting in the incorrect exact matching. Hence, fuzzy search is required, for which an LLM is used.

%To exploit this advantage of LLMs, we provided the entire list of existing instance names as inputs along with the instance to be matched. In the instruction, we ask the LLM to find the names on the instance name list that best matched the input instance. An exception ``None'' is also given as an option for those movies that are not on the list.

\smallskip\noindent
\textbf{Extra rules generation}. When humans chat socially, topic changes also happen when they find some common features of two topics involved. For example, one person might mention that in the movie Titanic, Jack sacrifices himself to save Rose, the listener may recall Romeo and Juliet, where sacrifice also happens between the lovers. This is also simulated by our AutoCompanion through LLMs' extra rules generation step with instructions given as:

{\small\tt In the [topic] [topic\_instance], [detailed\_content]. Is there any other [topic] that contains a similar [property]?}

\noindent Then, the list of topic instances and two examples are given to limit the scope of topic instances and control the output format. Table~\ref{tab:titanic} displays the result for the Titanic example from GPT-4, which is then added to the RCC list.

\begin{table*}[tb]
    \centering
    \caption{Output of GPT-4 of the similar plot to Titanic's sacrifice.}
    \label{tab:titanic}
    \medskip
%    \resizebox{\linewidth}{!}{
      \begin{minipage}{.9\linewidth}
        \begin{tabular}{p{3cm}p{8cm}}
          \toprule
          Movie Name & Similar Content\\
          \midrule
          The Dark Knight Rises & Batman sacrifices himself to save Gotham City, taking the blame for Harvey Dent's crimes and going into hiding.\\[.4em]

          The Hunger Games: Mockingjay - Part 2 &  Finnick sacrifices himself to allow Katniss and others to escape from mutts during the assault on the Capitol.\\
          \toprule
        \end{tabular}
      \end{minipage}
%    }
\end{table*}

\section{Conversation Samples}\label{sec:sample}

Below we display one conversation between the AutoCompanion and a human user, which demonstrates how the AutoCompanion directs and switches between topics.

\begin{description}[align=right,labelindent=!,leftmargin=1cm]
%  \addtolength{\itemsep}{-0.25\baselineskip}
  \tt \small
\item[Bot:] Hello! It's delightful coming across another cinema enthusiast. Seen any noteworthy films lately? Or maybe you've been immersed in a captivating read? My enthusiasm seems to oscillate between both constantly!
\item[User:] Me too! I just saw Inception. It is a great idea to take action on one's dream! Dreams in the dreams! What a fabulous idea!
\end{description}
\vspace{-0.25in}
\begin{description}[align=right,labelindent=!,leftmargin=1cm]
%  \addtolength{\itemsep}{-0.25\baselineskip}
  \tt \small
\item[Themes:]
\end{description}
\vspace{-0.15in}
\begin{lstlisting}[style=Themes]
talk(movie, Inception, plot episode). content(plot episode, actions in dreams). attitude(positive).
\end{lstlisting}
\vspace{-0.15in}
\begin{description}[align=right,labelindent=!,leftmargin=1cm]
%  \addtolength{\itemsep}{-0.25\baselineskip}
  \tt \small
\item[Next:]
\end{description}
\vspace{-0.15in}
\begin{lstlisting}[style=Themes]
talk(movie,Inception,plot episode). attitude(positive).
\end{lstlisting}
\vspace{-0.1in}
\begin{description}[align=right,labelindent=!,leftmargin=1cm]
%  \addtolength{\itemsep}{-0.25\baselineskip}
  \tt \small
\item[Bot:] Absolutely! The concept of a dream within a dream in Inception is simply astonishing. Who could forget that sensational zero-gravity hotel scene? The representation of physics and action was remarkable, wasn't it?
\item[User:] Yeah! And those people are kicked to wake up from their level of dreams one after another. very impressive and surprising, right?
\end{description}
\vspace{-0.25in}
\begin{description}[align=right,labelindent=!,leftmargin=1cm]
%  \addtolength{\itemsep}{-0.25\baselineskip}
  \tt \small
\item[Themes:]
\end{description}
\vspace{-0.15in}
\begin{lstlisting}[style=Themes]
talk(movie, Inception, plot episode). content(plot episode, waking up one after another). attitude(positive).
\end{lstlisting}
\vspace{-0.15in}
\begin{description}[align=right,labelindent=!,leftmargin=1cm]
%  \addtolength{\itemsep}{-0.25\baselineskip}
  \tt \small
\item[Next:]
\end{description}
\vspace{-0.15in}
\begin{lstlisting}[style=Themes]
talk(movie,The Wolf of Wall Street,plot episode). attitude(positive).
\end{lstlisting}
\vspace{-0.1in}
\begin{description}[align=right,labelindent=!,leftmargin=1cm]
%  \addtolength{\itemsep}{-0.25\baselineskip}
  \tt \small
\item[Bot:] Totally! Your mention of 'Inception' brings to mind 'The Wolf of Wall Street', another DiCaprio masterpiece. The scene in which Jordan Belfort galvanizes his firm and denounces stepping down is incredibly grand! Quite the adrenaline surge, right?
\item[User:]  Sorry I haven't seen this movie before. But I did see another of his work, Catch Me If You Can. I think DiCaprio's acting there quite matches the character's traits, right? It's really a good story. I like that.
\end{description}
\vspace{-0.25in}
\begin{description}[align=right,labelindent=!,leftmargin=1cm]
%  \addtolength{\itemsep}{-0.25\baselineskip}
  \tt \small
\item[Themes:]
\end{description}
\vspace{-0.15in}
\begin{lstlisting}[style=Themes]
talk(person, Leonardo DiCaprio, filmography). content(filmography, Catch Me If You Can). attitude(positive).
talk(movie, Catch Me If You Can, actor performance). content(actor performance, acting of DiCaprio matches the traits). attitude(positive).
talk(movie, Catch Me If You Can, plot episode). attitude(positive).
\end{lstlisting}
\vspace{-0.15in}
\begin{description}[align=right,labelindent=!,leftmargin=1cm]
%  \addtolength{\itemsep}{-0.25\baselineskip}
  \tt \small
\item[Next:]
\end{description}
\vspace{-0.15in}
\begin{lstlisting}[style=Themes]
talk(movie,Catch Me If You Can,plot episode). attitude(positive).
\end{lstlisting}
\vspace{-0.1in}
\begin{description}[align=right,labelindent=!,leftmargin=1cm]
%  \addtolength{\itemsep}{-0.25\baselineskip}
  \tt \small
\item[Bot:]  Couldn't agree more, buddy - the scene with Frank as a Pan Am pilot? Unbelievably good. It's astounding to think of a kid being able to pull off a scam that big. What scene did you enjoy the most?
\item[User:]  Right, that's amazing! But almost everybody trusts his make-up identity without a second thought. It's kind of ridiculous, but if things like this happened around us, it would be terrible, right?
\end{description}
\vspace{-0.25in}
\begin{description}[align=right,labelindent=!,leftmargin=1cm]
%  \addtolength{\itemsep}{-0.25\baselineskip}
  \tt \small
\item[Themes:]
\end{description}
\vspace{-0.15in}
\begin{lstlisting}[style=Themes]
talk(movie, Catch Me If You Can, characterization). content(characterization, everybody trusts Frank's make-up identity). attitude(negative).
talk(movie, Catch Me If You Can, social impact). content(social impact, terrible if happened in real life). attitude(positive).
\end{lstlisting}
\vspace{-0.15in}
\begin{description}[align=right,labelindent=!,leftmargin=1cm]
%  \addtolength{\itemsep}{-0.25\baselineskip}
  \tt \small
\item[Next:]
\end{description}
\vspace{-0.15in}
\begin{lstlisting}[style=Themes]
talk(movie,Catch Me If You Can,social impact). attitude(positive).
\end{lstlisting}
\vspace{-0.1in}
\begin{description}[align=right,labelindent=!,leftmargin=1cm]
%  \addtolength{\itemsep}{-0.25\baselineskip}
  \tt \small 
\item[Bot:] Absolutely, buddy! Imagine having tricksters like Frank amidst us - entertaining in films but not in reality! The movie, Catch Me If You Can, prompts us not to be rebels but to demand transparency, wouldn't you agree? There's a certain allure to the con artist lifestyle, don't you find it intriguing?
\item[User:]   So according to you, this is a very educational movie, right? Anyway I really like its story, it's fun and exciting.
\end{description}
\vspace{-0.25in}
\begin{description}[align=right,labelindent=!,leftmargin=1cm]
%  \addtolength{\itemsep}{-0.25\baselineskip}
  \tt \small
\item[Themes:]
\end{description}
\vspace{-0.15in}
\begin{lstlisting}[style=Themes]
talk(movie, Catch Me If You Can, value expressed). content(value expressed, educational). attitude(positive).
talk(movie, Catch Me If You Can, plot episode). content(plot episode, fun and exciting). attitude(positive).
\end{lstlisting}
\vspace{-0.15in}
\begin{description}[align=right,labelindent=!,leftmargin=1cm]
%  \addtolength{\itemsep}{-0.25\baselineskip}
  \tt \small
\item[Next:]
\end{description}
\vspace{-0.15in}
\begin{lstlisting}[style=Themes]
talk(movie,Don't Look Up,plot episode). attitude(positive).
\end{lstlisting}
\vspace{-0.1in}
\begin{description}[align=right,labelindent=!,leftmargin=1cm]
%  \addtolength{\itemsep}{-0.25\baselineskip}
  \tt \small
\item[Bot:] I totally understand, it's an entertaining narrative full of excitement, with underlying themes of trust and integrity. Your mention of Catch Me If You Can brings to mind another film featuring Leonardo DiCaprio, namely Don't Look Up. Well, personal opinion aside, the sequence with the meteorite approaching Earth was particularly gripping. Would you like to share your thoughts on it?
\item[User:]  Ah I don't quite like that movie. It thinks it satirizes a lot of things, but there is nothing fresh or original. It is neither spicy nor funny, and its reflection on the political situation at that time is a bit deliberate.
\end{description}
\vspace{-0.25in}
\begin{description}[align=right,labelindent=!,leftmargin=1cm]
%  \addtolength{\itemsep}{-0.25\baselineskip}
  \tt \small
\item[Themes:]
\end{description}
\vspace{-0.15in}
\begin{lstlisting}[style=Themes]
talk(movie, Don't Look Up, plot episode). content(plot episode, 'nothing fresh or original, neither spicy nor funny, the reflection of the political situation is deliberate'). attitude(negative).
\end{lstlisting}
\vspace{-0.15in}
\begin{description}[align=right,labelindent=!,leftmargin=1cm]
%  \addtolength{\itemsep}{-0.25\baselineskip}
  \tt \small
\item[Next:]
\end{description}
\vspace{-0.15in}
\begin{lstlisting}[style=Themes]
talk(person,Jennifer Lawrence,filmography). attitude(negative).
\end{lstlisting}
\vspace{-0.1in}
\begin{description}[align=right,labelindent=!,leftmargin=1cm]
%  \addtolength{\itemsep}{-0.25\baselineskip}
  \tt \small
\item[Bot:]  Absolutely, my friend. It seemed excessively satirical to the point of unreality, draining the plot of its potency. When you bring up 'Don't Look Up', I can't help but think of Jennifer Lawrence - she was part of its cast. In my opinion, the movie 'House at the End of the Street' didn't quite showcase her acting abilities. A talent like hers seemed underutilized, don't you agree?
\item[User:] I think it's the role that limits her, although in fact her character is actually one of the few bright spots in this movie.
\end{description}
\vspace{-0.25in}
\begin{description}[align=right,labelindent=!,leftmargin=1cm]
%  \addtolength{\itemsep}{-0.25\baselineskip}
  \tt \small
\item[Themes:]
\end{description}
\vspace{-0.15in}
\begin{lstlisting}[style=Themes]
talk(person, Jennifer Lawrence, acting skill). content(acting skill, limited by role in House at the End of the Street). attitude(negative).
talk(movie, House at the End of the Street, actor performance). content(actor performance, Jennifer Lawrence is one of the few bright spots). attitude(positive).
\end{lstlisting}
\vspace{-0.15in}
\begin{description}[align=right,labelindent=!,leftmargin=1cm]
%  \addtolength{\itemsep}{-0.25\baselineskip}
  \tt \small
\item[Next:]
\end{description}
\vspace{-0.15in}
\begin{lstlisting}[style=Themes]
talk(movie,House at the End of the Street,actor performance). attitude(positive).
\end{lstlisting}
\vspace{-0.1in}
\begin{description}[align=right,labelindent=!,leftmargin=1cm]
%  \addtolength{\itemsep}{-0.25\baselineskip}
  \tt \small
  \item[Bot:] Absolutely, I concur with your point. Jennifer Lawrence's skills exceeded her role. Her performance was indeed the bright spot in the film. The fear and struggle she represented was so real. Do you have any insights regarding the performance of other actors?
  \item[User:] The male lead is quite handsome, and the ending is really powerful, adding a lot of color to the movie.
\end{description}
\vspace{-0.25in}
\begin{description}[align=right,labelindent=!,leftmargin=1cm]
%  \addtolength{\itemsep}{-0.25\baselineskip}
  \tt \small
\item[Themes:]
\end{description}
\vspace{-0.15in}
\begin{lstlisting}[style=Themes]
talk(movie, House at the End of the Street, actor performance). content(actor performance, male lead is handsome). attitude(positive). 
talk(movie, House at the End of the Street, plot episode). content(plot episode, powerful ending). attitude(positive).
\end{lstlisting}
\vspace{-0.15in}
\begin{description}[align=right,labelindent=!,leftmargin=1cm]
%  \addtolength{\itemsep}{-0.25\baselineskip}
  \tt \small
\item[Next:]
\end{description}
\vspace{-0.15in}
\begin{lstlisting}[style=Themes]
talk(movie,House at the End of the Street,plot episode). attitude(positive).
\end{lstlisting}
\vspace{-0.1in}
\begin{description}[align=right,labelindent=!,leftmargin=1cm]
%  \addtolength{\itemsep}{-0.25\baselineskip}
  \tt \small
  \item[Bot:] Indeed, the finale, oh my, totally transformed the game! It added an exciting edge to the film! Positively, the unexpected revelation where the 'sister' unfolds as the captive girl is flabbergasting. I was completely caught off guard. Did this astonish you to the same degree as it did me?
  \item[User:] Yeah! But simply astonished. Nothing else. I'm happy to talk with you, but I need to go now. See you next time!
\end{description}
\vspace{-0.25in}
\begin{description}[align=right,labelindent=!,leftmargin=1cm]
%  \addtolength{\itemsep}{-0.25\baselineskip}
  \tt \small
\item[Themes:]
\end{description}
\vspace{-0.15in}
\begin{lstlisting}[style=Themes]
talk(movie, House at the End of the Street, emotion impact). content(emotion impact, just astunished). attitude(negative).
quit.
\end{lstlisting}
\vspace{-0.15in}
\begin{description}[align=right,labelindent=!,leftmargin=1cm]
%  \addtolength{\itemsep}{-0.25\baselineskip}
  \tt \small
\item[Next:]
\end{description}
\vspace{-0.15in}
\begin{lstlisting}[style=Themes]
quit.
\end{lstlisting}
\vspace{-0.1in}
\begin{description}[align=right,labelindent=!,leftmargin=1cm]
%  \addtolength{\itemsep}{-0.25\baselineskip}
  \tt \small
  \item[Bot:] Sure thing! I had a great time talking about movies and books with you. I hope we can carry on with our conversation next time. Have a great day!
\end{description}

\section{Evaluation}

Next, we evaluate AutoCompanion's execution efficiency and quality of dialog generation.

\begin{table}[tb]
    \centering
    \caption{Time cost for s(CASP) reasoning call and total response generation.}
    \label{tab:time}
    \medskip
 %   \resizebox{\linewidth}{!}{
\begin{minipage}{.7\linewidth}
    \begin{tabular}{lr}
 %   \centering
    \toprule
       Task  & Time Cost for Execution\\
    \midrule
        s(CASP) Reasoning (for one call) & 8ms\\

        Total (for each round) &  11440ms\\
    \toprule
    \end{tabular}
    \end{minipage}
%    }
\end{table}

\smallskip\noindent\textbf{Execution Time.}
We experimented with the processing time for each round of reply generation in Table \ref{tab:time}, and the average time consumption for 50 rounds is 11.44s, which is an acceptable waiting time for social sessions. The result also shows that the average time consumption for ASP reasoning is about 8ms, indicating that most of the time is spent in GPT-4 call. These experiments are done with an Intel i7-9800X CPU @ 3.80GHz and 32 GB RAM. We believe that LLMs with faster response times will emerge shortly, and this cost of time will be further shortened. Also, having the LLM reside locally on the machine will speed up response time.

%GG: We should mention what kind of machine was used for this measurement

% reviewer update 
\smallskip\noindent\textbf{Dialog Quality.}
We employed GPT-4 as a judge to evaluate the conversation quality following the instruction of \cite{judge}. AutoCompanion is compared with ChatGPT-3.5 in aspects of creativity, topic concentration, and conversation depth. We provided three pieces of conversation each, where the first one starts with the book \textit{The Little Prince}, the second one initiates with the movie \textit{Titanic}, and \textit{The Lord of Rings} by J.R.R. Tolkien for the third one. For each aspect, we provided the two conversations from different bots and asked the GPT-4 judge to distinguish which one was better. Due to the position-sensitivity of GPT-4 reported by \cite{judge}, for each experiment we asked twice, where the order of the conversations was swapped. The experiment result is shown in Table \ref{tab:eval}, indicating that AutoCompanion performs better than ChatGPT-3.5 in creativity and topic concentration, which is consistent with our observations. A more comprehensive evaluation is part of our future work. 

\begin{table}[tb]
  % \ref{table-comp}
    \centering
    \caption{We compared AutoCompanion with ChatGPT-3.5 on creativity, topic concentration, and conversation depth by employing an LLM-as-a-Judge system. The result shows that AutoCompanion outperforms ChatGPT-3.5 in creativity and has better control of the topic.}
    \label{tab:eval}
    \medskip
 %   \resizebox{\linewidth}{!}{
\begin{minipage}{.8\linewidth}
    \begin{tabular}{l|rrr}
 %   \centering
    \toprule
       Conversation  & Creativity & Topic Concentration & Conversation Depth\\
    \midrule
        Conv 1 & \textbf{AutoCompanion} & tie & tie \\

        Conv 2 &  \textbf{AutoCompanion} & tie & tie \\

        Conv 3 & tie & \textbf{AutoCompanion} & tie\\
    \toprule
    \end{tabular}
    \end{minipage}
%    }
\end{table}

\section{Discussion}\label{sec:discussion}

In this section, we discuss the features of AutoCompanion and the proper ways to use LLMs.

\smallskip\noindent\textbf{Why is AutoCompanion better than the LLM-only socialbots?}
Compared with LLM-only socialbot, our AutoCompanion has the following advantages: (i) Our AutoCompanion is based on the knowledge base, not pattern matching alone, which allows retrieval of relevant information about any topic instance through a query, avoiding factual errors. 
% reviewer update 9
When we only use LLM to build socialbots, since the knowledge is not explicitly stored in a specific library, but is compressed and stored using neural networks and weights, this will inevitably lead to confusion between concepts, and hallucinations will emerge. 
For instance, when asking GPT-4 the below question:

{\tt\small Is Currie Graham in the movie Hitchcock?}

\noindent It would reply:

{\tt\small Yes, Currie Graham is in the movie "Hitchcock." He plays the role of Bertie, one of the characters involved in the production of "Psycho."}

\noindent However, according to IMDb, Currie Graham acts as a PR called Flack in this movie and thus can be corrected through AutoCompanion. There are other advantages of AutoCompanion:

\begin{itemize} 
\item The use of an information database enhances the scalability of AutoCompanion. To update an LLM, we need to either load extremely long relevant texts at any time or train the LLM for a long time. But if a database supplements our information, an update to the knowledge base is all that is needed.

\item ASP reasoner's ability to control LLM is also a major advantage. LLMs usually require to describe the instructions in as much detail as possible to complete the task better (change topic, recommend, etc.).
% reviewer update 9
In the socialbot task for LLMs, this is reflected in the limitation of conversation topics, the choice of strategies for talking about a certain topic for several rounds and then jumping to another topic, the choice of recommendation timing for the latest popular topic instances, etc. 
However, these instructions usually require a large amount of token space, which will gradually be ignored as the conversation length increases. On the other hand, we don't record the specific content of the conversation but the state is saved in the summary.
% AR - Commenting below line, unclear.
% it can neither take advantage of LLM in capturing contextual implicit knowledge, nor is it as reliable in reasoning as ASP reasoner.

\item The ASP reasoner can specify some less common, out-of-the-box-thinking 
%generation directions 
for an LLM. Even with a certain degree of randomness, it is difficult for LLMs to truly generate some opinions that are far away from most people's thinking. Yet, %each individual's preferences are different. 
even if a movie has excellent reviews, there remains a small group of people who don't like it for various reasons. %When we socialize with humans, even if the probability of encountering such a point of view is extremely low, we can always encounter some due to the wide range of topics discussed. 
However, when using an LLM directly, it almost always caters to the mainstream views of the public and lacks its ``own unique insights''. This can also be avoided by cleverly using an ASP reasoner. %(e.g., specify the opinion generated is positive/negative). 
%For example, the reasoner directly specifies whether the generated topic is positive/negative so that LLM can generate some niche but unique perspectives.
\end{itemize}

\smallskip\noindent
\textbf{Is it proper to let LLM generate the fact-based content for use by AutoCompanion?}
It is well known that LLMs are good at generating coherent text. However, the accuracy of the generated content is not guaranteed. If we have certain requirements on the correctness of the generated content, relying only on an LLM may not be a good choice. 
Nonetheless, with the increase in training materials, the reliability of LLM has also improved to a certain extent. 
%For example, GPT-4 can accurately point out factual errors in what a user says, even if the user asks the question in a misleading tone. If you ask ``Why did William Shakespeare kill himself?'', GPT-4 will recognize this as a trap and say ``Oh, well you seem to have been misinformed. William Shakespeare, the famous English playwright, didn't commit suicide.'' 
%However, this is not the case for every response: if asked about an unfamiliar field, such as a recently released movie, or a book few people have read, hallucination remains.
%
In such a case then, would it be better to rely only on the knowledge in the knowledge base, and allow LLM to only paraphrase? Not necessarily, because this requires an extremely detailed and complete knowledge base, which we may not have,
% reviewer update 9
such as storing all the information about each movie, including all plots, shots, lines, costumes, sets, and other information, as well as opinions about them, and this is almost impossible. 
For our AutoCompanion, since almost all movies/books/people are well-known to the public, GPT-4 has also been exposed to relevant materials and is familiar with relevant information, so it is feasible for GPT-4 to generate content directly, but it also sacrifices some accuracy. 
% reviewer update 10
And, since enough constraints are imposed and the sentences are short, we have not observed hallucinated responses in our reasonably extensive experiments.
In our future work, the knowledge base will be expanded to cover most movies and books, but if so, a detailed knowledge base is required.
%Therefore, we need to make a trade-off between reliability and availability.
%If a huge knowledge base is required to cover most movies and books, it should be compromised to only record partial information. And for those with knowledge out of the base, it can simply evade by saying that it does not know. 
%Because ordinary people, if they are not ardent fans, will not carefully appreciate every detail of a movie or a book.

%GG: should we comment out the text below? Seems irrelevant
%We can also allow LLMs access to the internet. 
%%LLM can capture the latest and most detailed knowledge by turning to the Internet. 
%However, the information on the Internet has not been screened, so its reliability cannot be guaranteed either.

% reviewer update 4
\smallskip\noindent 
\textbf{Why does LLM always generate answers lacking novelty?}
\label{sec:boring}
When we used GPT-4 to build and use an LLM-only socialbot, we found that the views generated by GPT-4 were often old and boring, not only wrt the property of a certain topic instance, but also wrt the association of one instance to another. For example, when discussing fantasy literature, GPT-4 will always mention ``Middle-Earth'' or the ``Harry Potter'' series first and express its appreciation for them. As another example, when we mention the movie Inception or other acclaimed movies, GPT-4 always speaks highly of it. Even when we ask it to talk about its shortcomings about a certain property, it sometimes replies, ``I can't think of any shortcomings.'' Work of \cite{llm_bias} suggests that this is related to the bias and fairness of LLM to a certain extent. Since the data injected into LLM is based on statistics, LLM's belief in a certain point of view is constantly strengthened during the training. Especially when a small number of samples are carefully labeled but used repeatedly, those data that deviate from the general public opinion are directly eliminated. At the same time, training to align reinforcement learning with users also weakens the creativity of LLM to a certain extent. This essentially reflects the ``data organizer'' attribute of LLM, that is, the current LLM is only trained for data collection and organization, and is rarely trained for innovative work -- and innovation is difficult to gain from training. \cite{creativity} also pointed out that although LLM can produce texts of a certain value, revolutionary or eye-catching ideas are rarely found.

\section{Conclusion and Future Work}

This paper presents an approach to building a socialbot that utilizes LLM to translate between natural language and predicates while employing an ASP reasoner as the backend. 
%This approach succeeds in making the reasoning system understand the user's natural language input and mitigates the unpredictability associated with LLM. 
%In the reasoner stage, we introduced two technologies, the Conversational Knowledge Template and Relevant Consistent Concept, and built a reasoning system based on them that integrates deeper topics, initiating topics, answering questions, and recommending new topic instances. 
We used this framework to develop the AutoCompanion socialbot to chat about movies and book-related topics. AutoCompanion outperforms LLM-only socialbots in reliability, scalability, ease of use, and creativity. We believe this organic combination of LLM and ASP is one of the most effective ways to realize the control of LLM-based socialbots.

Our future work includes building a more complete knowledge base that allows socialbots to query most of the topic knowledge rather than generating it using an LLM; it also includes field testing and evaluation of socialbots. We expect to make our AutoCompanion Socialbot available on online platforms shortly. Feedback from users will be one of the criteria to measure the performance of our AutoCompanion. 
% reviewer update 6
Additionally, continuing to explore how to better utilize LLMs to generate unstructured knowledge and ensure its reliability 
%(probably by RAG) 
is also part of our future work. We are also planning to add multi-modality, like importing movie clips and voice I/O to our AutoCompanion.
%\vspace{-0.2in}

\section*{Acknowledgement}

Authors have been partially supported by US NSF Grants IIS 1910131, US DoD, grants from industry through the UT Dallas Center for Applied AI and Machine Learning, and grants VAE (TED2021-131295B-C33) and COSASS (PID2021-123673OB-C32) funded by MCIN/AEI. The authors would like to thank the anonmyous reviewers for their insightful comments and suggestions for improvement.

\bibliographystyle{tlplike}
\bibliography{generic}

\end{document}